\documentclass[10pt,twocolumn,letterpaper]{article}

\usepackage{iccv}
\usepackage{times}
\usepackage{epsfig}
\usepackage{graphicx}
\usepackage{amsmath}
\usepackage{amssymb}
\usepackage{subfigure}
\usepackage{algorithmic}
\usepackage[algoruled,vlined]{algorithm2e}

\usepackage[pagebackref=true,breaklinks=true,letterpaper=true,colorlinks,bookmarks=false]{hyperref}

\iccvfinalcopy % *** Uncomment this line for the final submission

\def\iccvPaperID{27} % *** Enter the ICCV Paper ID here

\ificcvfinal\fi
\begin{document}

%%%%%%%%% TITLE
\title{Learning Structured Ordinal Measures for Video based Face Recognition}

\author{Ran He$^a$, Tieniu Tan$^a$, Larry Davis$^b$, and Zhenan Sun$^a$\\
$^a$Institute of Automation, Chinese Academy of Science. Beijing, China 100190.\\
$^b$University of Maryland. College Park, MD, USA 20742.\\
{\tt\small \{rhe,tnt,znsun\}@nlpr.ia.ac.cn, lsd@umiacs.umd.edu}}
% For a paper whose authors are all at the same institution,
% omit the following lines up until the closing ``}''.
% Additional authors and addresses can be added with ``\and'',
% just like the second author.
% To save space, use either the email address or home page, not both
%\and
%Larry Davis\\
%University of Maryland\\
%College Park, MD, USA 20742\\
%{\tt\small lsd@umiacs.umd.edu}
%}

\maketitle
%\thispagestyle{empty}

%%%%%%%%% ABSTRACT
\begin{abstract}
   This paper presents a structured ordinal measure method for video-based face recognition that simultaneously learns ordinal filters and structured ordinal features. The problem is posed as a non-convex integer program problem that includes two parts. The first part learns stable ordinal filters to project video data into a large-margin ordinal space. The second seeks self-correcting and discrete codes by balancing the projected data and a rank-one ordinal matrix in a structured low-rank way. Unsupervised and supervised structures are considered for the ordinal matrix. In addition, as a complement to hierarchical structures, deep feature representations are integrated into our method to enhance coding stability. An alternating minimization method is employed to handle the discrete and low-rank constraints, yielding high-quality codes that capture prior structures well. Experimental results on three commonly used face video databases show that our method with a simple voting classifier can achieve state-of-the-art recognition rates using fewer features and samples.
\end{abstract}

%%%%%%%%% BODY TEXT
\section{Introduction}
Video-sharing websites are a fast-growing platform that allows internet users to distribute their video clips. There are often a large number of face videos in these websites. How to index, retrieve, and classify these face videos has become an active research topic in the area of video-based face recognition (VFR). Current VFR methods often perform recognition based on hundreds or thousands of floating point features, and store almost every face sample from a video clip. Since there can be (many) thousands of face samples in a video clip, high-dimensional dense features and large-scale registered samples result in tremendously large time and space complexity, which becomes a computational bottleneck when applying VFR methods to video-sharing websites.

Recently, binary code representations have drawn much attention in biometric recognition \cite{ZChai:2014}\cite{ZLei:2014}\cite{JLu:2015} and large scale image retrieval~\cite{WLiu:2012}\cite{Grauman:2013}\cite{GLin:2014}. Among these binary coding methods, codes constructed from ordinal measures (OM) are one representative method. Ordinal measures \cite{Stevens:1946} are common in human perceptual judgments. It is easy and natural for humans to rank or order the heights of two persons, although it is hard to estimate their precise differences~ \cite{ZSun:2014}. Ordinal measures were originally used in social science \cite{Stevens:1946} and then introduced to computer vision.

\begin{figure*}[t]
\centering
\includegraphics[width=132mm]{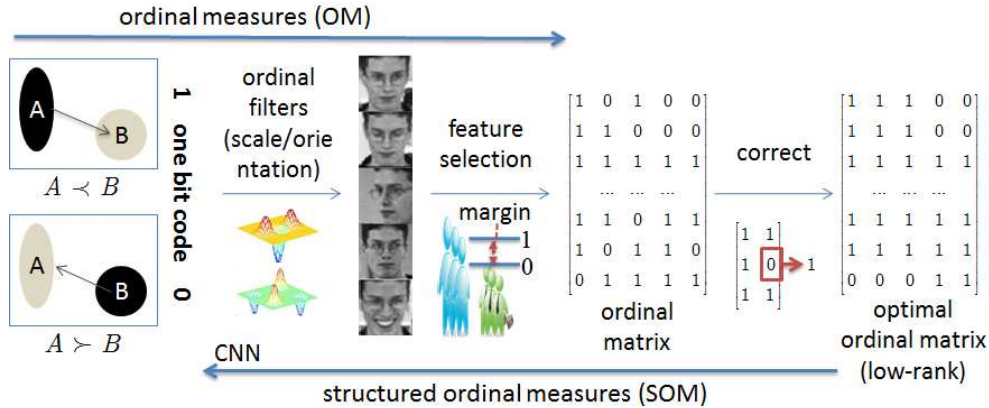}
\caption{An illustration of structured ordinal measures. Ordinal measure of visual relationship between two regions \cite{ZSun:2009}\cite{ZSun:2014}. Previous OM methods apply feature selection methods to select over-complete ordinal features (binary codes) that are generated by handcrafted ordinal filters. SOM simultaneously seeks ordinal filters and optimal ordinal features in a data-driven way, makes the learned features low-rank and enforces an optimal ordinal matrix for classification. In SOM, one binary code of a sample can be corrected according to the codes of similar samples. \label{fig:spot}}
\end{figure*}

In biometrics, an OM is defined as the relative ordering of some property - for example, the average brightness of two adjacent regions (with 1 coding $A \succ B$ and 0 coding $A \prec B$) or the relative ordering of two color channels within the same region. Ordinal filters with a number of tunable parameters, are methods to analyze the ordinal measures of image features. The Haar wavelet and quadratic spline wavelet can be regarded as typical ordinal filters. Ordinal features are the binary codes of image features obtained by thresholding ordinal filters. Fig.~\ref{fig:spot} plots a simple illustration of OM.

In prior work, the set of handcrafted ordinal filters is chosen to correspond to some family of coherent patterns - like Gabor filters.  The space of ordinal filters can therefore be quite large as the tunable parameters - scale, frequency, orientation - are varied, each giving rise to a potential ordinal feature. Different feature selection methods \cite{ZSun:2009}\cite{ZSun:2014}\cite{LXiao:2013} have been used for OM to select a stable subset from the over-complete ordinal features. The term 'stable' indicates that the floating point features generated by an ordinal filter from the same class are expected to have large margins so that the corresponding ordinal features (binary codes) are robust to intra-class variations during binarization.

%The first issue is the design of ordinal filters. The existing ordinal filters are often handcrafted. But handcrafted ordinal filters are too simple to represent complex human vision structures \cite{SLiao:2007}. More important, these filters are designed based on human experience and ignore useful information from data. In addition, to improve classification accuracy, these filters often contain a large number of parameters based on distance, scale and location, resulting in a potential feature set of OM's. This naturally leads to the second issue, i.e., how to select the optimal set of ordinal features. Although boosting~\cite{ZSun:2009}, sparse representation \cite{LXiao:2013} and linear programming~\cite{ZSun:2014} have been employed to improve selection results, it is still difficult for a feature selection algorithm to select the optimal set from the over-complete set of OM's.

Motivated by the success of OM in iris \cite{ZSun:2009}, palmprint~\cite{ZSun:2014} and face recognition \cite{ZChai:2014}, we present what we refer to as a structured ordinal measure (SOM) method for video-to-video face recognition. Different from previous handcrafted OM methods, SOM simultaneously learns ordinal filters (SVM's) and structured ordinal features (binary codes) from video data as shown in Fig.~\ref{fig:spot}. Considering that face appearances in video clips contain several facial variations and are similar in adjacent frames, we design the ordinal features of SOM to be stable and self-correcting binary codes. Stability indicates that the learned ordinal features are required to have large margins and to be clustered. The self-correcting character indicates that binary code of one frame depends not only on its corresponding ordinal filter (or coding function) but also on the binary values of similar (typically nearby in time) face samples. Because face images in a video clip often lie in a union of multiple linear subspaces~\cite{YChen:2013}\cite{GZhang:2014}, the features (binary code) assigned to the subset of faces from a  single linear subspace should be similar. These binary codes can be potentially corrected by each other through a low-rank constraint on the matrix of constructed codes. One of the main advantages of our method is that it simultaneously reduces the number of dense features and eliminates redundant samples\footnote{Getting rid of redundant samples is important during both training and testing. In a video clip, the face can remain unchanging for long periods of time and that would bias the models towards that appearance.}.

We will formulate the SOM problem as a non-convex integer program problem that mainly includes two parts. The first part learns stable ordinal filters to project video data into a space in which the filtered data are separable with a maximum margin. This can be viewed as an instance of maximum margin clustering (MMC)~\cite{LXu:2004}. The second finds self-correcting binary codes by balancing the projected real-value data and a rank-one ordinal matrix in a structured low-rank way. Unsupervised and supervised structures are considered for the ordinal matrix. We also integrate CNN feature representations into our method to enhance stability. An alternating optimization method provides an efficient discrete solution to deal with the discrete and low-rank constraints imposed on binary ordinal features. In addition, a simple voting classifier with a self-correcting process is proposed to efficiently compress and classify video clips. Experimental results on three commonly used face video databases show that our SOM method can achieve state-of-the-art recognition results using fewer features and samples. Compared to previous binary coding methods for still images (face or iris), SOM more efficiently utilizes the low-rank property of video data and hence is potentially useful for VFR problems.

There are three major contributions of this work:

1) By employing the optimal ordinal matrices as output structures, SOM encourages ordinal features from the same class to have similar binary codes. To the best of our knowledge, SOM is the first algorithm that learns binary codes (or hashing) using output structures.

2) Assuming that face images of a video clip lie in a union of linear subspaces, we propose a self-correcting method to discretely binarize both gallery and probe videos. Our method utilizes the continuous information in videos and hence is effective for VFR tasks.

3) As a by-product of SOM, we show that using a simple voting classifier improves over competing and complex classification models on fine grained datasets like the YouTube Celebrities dataset and offers an impressive compression ratio of CNN floating point features (20\% face samples and 64-bit binary codes).

The rest of this paper is organized as follows. We briefly review some recent advances on binary coding methods in Section~\ref{sec:rel}. In Section~\ref{sec:som} and Section~\ref{sec:opt}, we present the details of SOM and the optimal ordinal matrices respectively. Section~\ref{sec:exp} provides experimental results, prior to summary in Section~\ref{sec:con}.

\section{Related work \label{sec:rel}}
Since OM methods are an instance of binary appearance features, we briefly review some recent advances on binary coding methods.
\subsection{Biometric recognition}
In biometrics, binary feature representation methods often focus on directly computing local image patches by the filters to generate binary codes. Local binary patterns (LBP) and ordinal measures are two representative binary features. There are many variations of these two features~\cite{ZChai:2014}\cite{ZLei:2014}. The definition and properties of OM in the context of biometrics can be found in \cite{ZSun:2009}.

Although OM's has been successfully applied to biometrics, there are still two open issues for OM. The first issue is the design of ordinal filters. The existing ordinal filters are often handcrafted. But handcrafted ordinal filters are too simple to represent complex human vision structures \cite{SLiao:2007}. In addition, to improve stability and accuracy, these filters often contain a large number of parameters based on distance, scale and location, resulting in a potential feature set of OM. This naturally leads to the second issue, i.e., how to select the optimal set of ordinal features. Although various feature selection methods~\cite{ZSun:2009}\cite{LXiao:2013}\cite{ZSun:2014} have been employed to improve selection results, it is still difficult for a feature selection algorithm to select the optimal set from the over-complete set of OM.

%OM's are summarized as follows; {\bf Ordinal measures} are defined as the relative ordering of a set of regional image features (e.g. average intensity, Gabor wavelet coefficients, etc.). {\bf Ordinal filters} with a number of tunable parameters, are methods to analyze the ordinal measures of image features. The Haar wavelet and the derivative of Gaussian filter can be regarded as typical ordinal filters. {\bf Ordinal features} are the binary codes of image features learned by ordinal filters. Ordinal filters typically determine a huge set of ordinal features. Although OM's has been successfully applied to different biometric tasks\cite{ZSun:2009}\cite{ZSun:2014}\cite{ZChai:2014}, previous OM's completely depend on feature selection methods to reduce the redundancy of over-complete ordinal features \cite{ZSun:2009}\cite{ZSun:2014}. These feature selection methods often require a large number of ordinal features to obtain good accuracy and do not consider the output structure of the selected ordinal features.

Recently, data-driven binary feature methods, which learn local image filters from data, have drawn much attention. Cao et al.~\cite{ZCao:2010} utilized unsupervised methods (random-projection trees and PCA trees) to learn binary representations. Lei et al.~\cite{ZLei:2014} proposed a LBP-like discriminant face descriptor (DFD) by combining image filtering, pattern sampling and encoding. Chan et al.~\cite{TChan:2014} combined cascade PCA, binary code learning and block-wise histograms to learn a deep network. Lu et al.~\cite{JLu:2015} proposed a compact binary face descriptor (CBFD) to remove the redundancy information of face images. Although these methods indeed boost recognition performance on some challenging databases, their learned features are often high dimensional. For example, the dimensionality of histogram feature vectors of DFD and CBFD are 50,176 and 32,000 respectively. High dimensional and dense representations make these data-driven methods not applicable to VFR problems.

\subsection{Image retrieval}
Learning binary codes ('hashing') has been a key step to facilitate large-scale image retrieval. In image retrieval, the terminology 'hashing' refers to learning compact binary codes with Hamming distance computation. Similarity-sensitive hashing or locality-sensitive hashing algorithms~\cite{Weiss:2009}\cite{Kulis:2009_1}, graph-based hashing~\cite{WLiu:2014}, semi-supervised learning~\cite{JWang:2010_1}, support vector machine~\cite{YMu:2014}\cite{FShen:2015}, Riemannian manifold~\cite{YLi:2015}, decision trees~\cite{GLin:2014} and deep learning \cite{Grauman:2013}\cite{RXia:2014} have been studied to map high-dimensional data into a low-dimensional Hamming space. The authors in \cite{WLiu:2014}\cite{FShen:2015} argued that the degraded performance of hashing methods is due to the optimization procedures used to achieve discrete binary codes. Hence \cite{WLiu:2014}\cite{FShen:2015} tried to enforce binary constraints to directly obtain discrete codes~\cite{WLiu:2014}\cite{FShen:2015}. A brief review of hashing methods for image search can be found in \cite{Grauman:2013}\cite{JWang:2014}.

These hashing methods are often used for image search and retrieval but they may not achieve the highest accuracy for VFR problems. For example, the constraints in \cite{WLiu:2014} maximize the information from each binary code over all the samples in a training set. However, adjacent face samples in a video clip often have nearly the same appearance so that these samples can have similar binary codes. In addition, to the best of our knowledge, there is no existing hashing methods that address image-set problems~\cite{ZCui:2012}.

\section{Structured ordinal measures (SOM) \label{sec:som}}
\subsection{Motivation \label{sec:mot}}
Consider a training set $X$ from $C$ classes, which consists of $n$ biometric samples $x_j$ ($1 \le j \le n$) in a high dimensional Euclidean space $R^d$. The goal of previous OM methods is to identify ordinal filters over $X$ to nonlinearly map each $x_j$ to $m$ ordinal features (an m-bit binary code). Since ordinal filters typically have a number of tunable parameters and so determine a huge set of possible ordinal features, various feature selection methods have been used to select the $m$ ordinal features. The selected ordinal features of all samples form a binary matrix $B=[b_1,\dots, b_n] \in R^{m \times n}$, referred to as an {\bf ordinal matrix}. Previous OM methods select ordinal filters one by one (using a greedy approach) and hence neglect the output structure of ordinal features. For example, video data are often low-rank.

In biometrics, since intra-class variations of biometric samples are often very large, good ordinal measures should generate similar binary codes for the samples from one subject. In addition, a large difference between two quantities will result in more stable binary features. For example, the greater the color difference between two image regions, the more easily humans order their relative brightness (1 or 0); and the greater the height difference between two persons, the more easily humans rank their relative heights.

To obtain stable ordinal features, we introduce the following minimization problem for OM,
\begin{eqnarray}
&\mathop {\min }\limits_{W,\xi ,B} {\mu\xi  + \lambda _1 \left\| W \right\|_2  + \sum\limits_c {\left\| {B^c } \right\|_* } } \label{eq:obj}\\
&s.t. \; B_{ij} (w_i^T X_j ) \ge 1 - \xi _{ij}, \nonumber \\
& \quad\; \xi _{ij}  \ge 0, \; B_{ij}  \in \left\{ { - 1,1} \right\}\nonumber
\end{eqnarray}
where $\mu$ and $\lambda_1$ are constants, and $\left\| . \right\|_*$ denotes the matrix trace norm (i.e., the sum of its singular values). $B^c$ represents all ordinal features from the $c$-th class. The parameter matrix $W=[w_1,\dots, w_m] \in R^{d \times m}$ represents a set of ordinal filters. As defined in Section~\ref{sec:rel}, a parameter matrix $W$ contains a set of ordinal filters only if $W$ can result in consistent orders for the samples from the same class, e.g., $W^TX$ generates an ordinal matrix as in Fig~\ref{fig:stru}. In contrast to the binary coding methods~\cite{ZSun:2009}\cite{ZLei:2014}\cite{JLu:2015} that are based on local image patches, (\ref{eq:obj}) directly uses the whole image as an input to find compact codes\footnote{In face recognition, dividing a face image into small patches can capture nonlinear facial variations well and so improves recognition rates. The learned filters in (\ref{eq:obj}) can also be applied to local patches as in previous binary coding methods.}. More important, (\ref{eq:obj}) aims to simultaneously seek ordinal filters ($W$) and optimal ordinal features ($B$).

The low-rank constraint in (\ref{eq:obj}) encourages the ordinal features from the same class to be correlated. This constraint reduces the redundancy of video data and corrects some binary codes whose corresponding values ($W^TX$) are close to SVM's separating hyperplanes. We also want to enforce that the learned $B$ is close to the optimal ordinal (binary) matrix for classification, resulting in the following minimization problem,
\begin{eqnarray}\label{eq:obj_s}
&\mathop {\min }\limits_{W,\xi ,B} {\mu\xi  + \lambda _1 \left\| W \right\|_2  + \sum\limits_c {\left\| {B^c } \right\|_* } } + \lambda_2 \left\| {B - S} \right\|_F^2  \\
&s.t. \; B_{ij} (w_i^T X_j ) \ge 1 - \xi _{ij},  \xi _{ij}  \ge 0, \; B_{ij}  \in \left\{ { - 1,1} \right\}\nonumber
\end{eqnarray}
where $S \in R^{m \times n}$ is a prior ordinal matrix that defines a desired output structure for ordinal features. We postpone discussion of the design of $S$ until Section~\ref{sec:opt}. Since the OM problem in (\ref{eq:obj_s}) imposes an output structure on ordinal filter learning, we refer to the problem in (\ref{eq:obj_s}) as learning a {\bf structured ordinal measure}.

Even without the structured low-rank constraint, (\ref{eq:obj_s}) is difficult to solve~\cite{LXu:2004}. Unlike supervised SVM that can be formulated as a convex optimization problem, (\ref{eq:obj_s}), even without the structured low-rank constraint, is still a non-convex integer optimization problem. It is an instance of maximum margin clustering~\cite{LXu:2004}. To simplify the minimization of (\ref{eq:obj_s}), we relax (\ref{eq:obj_s}) by introducing an equality constraint on $B$ as follows,
\begin{eqnarray}
&\tiny{\mathop {\min } {\mu\xi  + \lambda _1 \left\| W \right\|_2  + \sum\limits_c {\left\| {B^c } \right\|_* } } + \lambda_2 \left\| {B - S} \right\|_F^2 + \left\| {E} \right\|_F^2 } \nonumber\\
&s.t. \; B = W^T X + E, \; B_{ij}  \in \left\{ { - 1,1} \right\}, \label{eq:obj_se}\\
&B_{ij} (w_i^T X_j ) \ge 1 - \xi _{ij},  \xi _{ij}  \ge 0 \nonumber
\end{eqnarray}
where $E \in R^{m \times n}$ is an error term to reduce the loss during binarization. Since $\left\| {B - S} \right\|_F^2  = \sum\nolimits_c {\left\| {B^c  - S^c } \right\|_F^2 }$, (\ref{eq:obj_se}) actually seeks discrete binary codes by balancing floating point data $W^T X$ and a rank-one ordinal matrix $S^c$ in a structured low-rank way.

Our SOM formulation in (\ref{eq:obj_se}) has two major advantages: 1) the introduction of the low-rank constraint and error term makes SOM more flexible during binarization. The learned binary codes depend on their corresponding floating point values as well as prior structures. Different from the binary codes that are directly generated by ordinal filters or hashing functions, the binary codes of SOM can be self-corrected by the structure constraints, resulting in self-correcting codes. 2) Since $S^c$ is a rank-one matrix, $\lambda_2$ plays the role of controlling the number of learning samples. The rank-one matrix indicates that there is only one unique sample in this matrix. The larger the value of $\lambda_2$, the more $B^c$ resembles $S^c$. In practice, the rank of $B^c$ will be larger than one because a face video clip often contains several face variations.

\subsection{Optimization \label{sec:sol}}
The optimization problem in (\ref{eq:obj_se}) is a hard computational problem (non-convex integer optimization), which belongs to the class of maximum margin clustering problems~\cite{LXu:2004}. Fortunately, we do not need to find the global minimum because local minima produce good ordinal features. Hence we can decompose the non-convex problem in (\ref{eq:obj_se}) into subproblems as in MMC. A local minimum can be obtained by solving a series of SVM training and binary code learning problems. An overview of our iterative algorithm is as follows.

First, fixing variables $B$ and $E$, we minimize (\ref{eq:obj_se}) w.r.t. variables $W$ and $\xi$, resulting in a multiple linear SVM problem in (\ref{eq:som_svm}) (one for each ordinal feature) \cite{RFan:2008}. To learn the $i$-th SVM \footnote{The $\ell_1$ regularized linear SVM is implemented by LIBLINEAR: \url{http://www.csie.ntu.edu.tw/~cjlin/libsvm}}, the columns of $X$ and the elements of the $i$th row of $B$ are used as training data and labels respectively.
\begin{eqnarray}
&\mathop {\min }\limits_{W,\xi}  \mu\xi  + \lambda _1 \left\| W \right\|_2  \label{eq:som_svm}\\
&s.t. \; B_{ij} (w_i^T X_j ) \ge 1 - \xi _{ij},  \xi _{ij}  \ge 0 \nonumber
\end{eqnarray}
Second, fixing variables $W$ and $\xi$, (\ref{eq:obj_se}) takes the following form w.r.t. $B$ and $E$,
\begin{eqnarray}
&\mathop {\min }\limits_{B,E} \sum\limits_c {\left\| {B^c } \right\|_* } + \lambda_2 \left\| {B - S} \right\|_F^2 + \left\| {E} \right\|_F^2 \label{eq:som_rank}\\
&s.t. \; B = A + E, \; B_{ij}  \in \left\{ { - 1,1} \right\} \nonumber
\end{eqnarray}
where $A=W^{\left\{t+1\right\}T} X$. By substituting the equality constraint into the objective function of (\ref{eq:som_rank}), we can reformulate (\ref{eq:som_rank}) as follows,
\begin{eqnarray}
&\mathop {\min }\limits_{B} \left\| {A-B} \right\|_F^2+\sum\limits_c {\left\| {B^c } \right\|_* } + \lambda_2 \left\| {B - S} \right\|_F^2 \label{eq:som_rank1}\\
&s.t. \; B_{ij}  \in \left\{ { - 1,1} \right\} \nonumber
\end{eqnarray}
Since $||.||_F^2$ is separable, the solution of (\ref{eq:som_rank1}) can be independently obtained by minimizing the following subproblem for each class $c$,
\begin{eqnarray}
&\mathop {\min }\limits_{B^c} \left\| {A^c-B^c} \right\|_F^2+{\left\| {B^c } \right\|_* } + \lambda_2 \left\| {B^c - S^c} \right\|_F^2 \label{eq:som_rank2}\\
&s.t. \; B^c_{ij}  \in \left\{ { - 1,1} \right\} \nonumber
\end{eqnarray}
To minimize the low-rank problem in (\ref{eq:som_rank2}), we first need to introduce a variational formulation for the trace norm \cite{Grave:2011},
\newtheorem{lemma}{Lemma}
\begin{lemma} \label{th:rank}
Let $B \in R^{m \times n}$. The trace norm of $B$ is equal to:
\begin{equation}
\left\| B \right\|_*  = {\textstyle{1 \over 2}}\mathop {\inf }\limits_{L \ge 0} tr\left( {B^T L^{ - 1} B} \right) + tr(L)
\end{equation}
and the infimum is attained for $L=(BB^T)^{1/2}$.
\end{lemma}
Using this lemma, we can reformulate (\ref{eq:som_rank2}) as,
\begin{eqnarray}
&\mathop {\min }\limits_{B^c} \mathop {\min }\limits_{L \ge 0}&{\left\| {A^c - B^c} \right\|_F^2  + tr(B^{cT} L^{ - 1} B^c)} \label{eq:som_rank3}\\
&&+ \lambda _2 \left\| {B^c - S^c} \right\|_F^2  + tr(L) \; s.t. \; B^c_{ij}  \in \left\{ { - 1,1} \right\} \nonumber
\end{eqnarray}
The problem in (\ref{eq:som_rank3}) can be alternately minimized. When $L$ is fixed, we can use the discrete cyclic coordinate descent method to obtain $B^c$ bit by bit. For simplicity, we develop a simple and direct method to find $B^c$. That is, disregarding the integer constraint, the solution of $B^c$ takes the following form by setting the derivative of (\ref{eq:som_rank3}) w.r.t. $B^c$ equal to zero,
\begin{equation}\label{eq:som_rankL}
B^c = ((1 + \lambda _2 )I + L^{ - 1} )\backslash (A^c + \lambda _2 S^c).
\end{equation}
Given a floating point $B^c$ in one iteration, we can use the sign function $sgn(.)$ to obtain binary-value $sgn(B^c)$. Experimental results show that the learned binary codes are good enough for VFR. Algorithm~\ref{Alg:SOM} summarizes the procedure to learn structured ordinal filters. $\lambda _2$ is set to 0.1 throughout this paper.

\begin{algorithm}[]
\caption{Learning structured ordinal filters \label{Alg:SOM}}
 \KwIn{Data matrix $X \in R^{d \times n}$ and ordinal matrix $S \in R^{m \times n}$}
 \KwOut{Ordinal Filters $W \in R^{d \times m}$}
\begin{algorithmic}[1]
    \REPEAT
    \STATE Train $m$ linear-SVMs to update $W$ using $B^{t-1}$ as training labels.
    \STATE Compute $A=\left\{W^t\right\}^TX$.
    \REPEAT
    \STATE Compute $L=(B^cB^{cT})^{1/2}$.
    \STATE Compute $B^c$ via (\ref{eq:som_rankL}).
    \STATE Let $B^c=sgn(B^c)$.
    \UNTIL The variation of $B$ is smaller than a threshold.
    \STATE t=t+1.
    \UNTIL The variation of $B$ is smaller than a threshold.
\end{algorithmic}
\end{algorithm}

\subsection{Classification \label{sec:cla}}
When applying SOM (or binary code learning methods) to biometric recognition, SOM must generate ordinal features for any data sample beyond the sample points in the training set $X$. Given a new probe dataset $X^p$, a hashing algorithm $H$ with parameter $W$ typically applies the sign function $sgn(.)$ to the hashing function $f_W^H (X^p)$ to obtain the binary codes~~\cite{WLiu:2014}\cite{FShen:2015}, i.e., $B^H = {\mathop{\rm sgn}} (f_W^H (X^p))$.

VFR can be viewed as an image-set classification/retrival problem~\cite{ZCui:2012}. The samples in a probe (or gallery) dataset are from a video clip and so have a low-rank structure. Hence, instead of using the sign function, we propose a low-rank method to construct the binary codes for a probe video as follows,
\begin{eqnarray}
&\mathop {\min }\limits_B \left\{ {\left\| E \right\|_F^2  + \left\| B \right\|_* } \right\} \label{eq:cor} \\
& s.t. \; {B=f_W^H (X^p)+E},\; B_{ij}  \in \left\{ { - 1,1} \right\} \nonumber
\end{eqnarray}
Compared to directly using the sign function $sgn(.)$ to obtain binary codes, (\ref{eq:cor}) utilizes a low-rank prior to find binary codes. This makes the binary codes $B$ not only depend on the function $f_W^H (.)$. The values in $B$ can be potentially changed (or corrected) by each other due to the low-rank constraint. (\ref{eq:cor}) is a sub-problem of (\ref{eq:som_rank2}) when $\lambda_2$ is set to zero. Hence (\ref{eq:cor}) can be alternatively minimized as (\ref{eq:som_rank2}).

Given the binary codes constructed from (\ref{eq:cor}), a simple nearest neighbor classifier for each unique code in $B$ (since many samples can be mapped to the same code by the optimization) with voting is used as classifier to report recognition rates. The class label of the majority class in a video sequence is taken as the final class label of this sequence. In addition, since the low-rank constraint in (\ref{eq:cor}) tends to make the column samples in $B$ correlated, it also tends to reduce the number of different samples in $B$. We introduce the term {\bf compression ratio of samples} for VFR, i.e., compression ratio = the number of unique samples/ the total number of samples. A lower compression ratio of an algorithm indicates that the algorithm needs less storage space (and as a consequence less computational time).

In addition, since there is no a rank-one constraint in (\ref{eq:cor}) (compared to (\ref{eq:obj_s})), compression ratio will tend to be high as the number of desired bits increases. If some priors of the rank of a video clip are given or a lower compression ratio is required, we can further impose a rank constraint on (\ref{eq:cor}), resulting in the following minimization problem,
\begin{eqnarray}
&\mathop {\min }\limits_B \left\| {f_W^H (X^p ) - B} \right\|_F^2 \label{eq:cor2}\\
& s.t. \; rank(B)\le r, \; B_{ij}  \in \left\{ { - 1,1} \right\} \nonumber
\end{eqnarray}
where $rank(.)$ is the matrix rank operator and $r$ is constant. The rank constraint in (\ref{eq:cor2}) makes the rank of $B$ is smaller than $r$. That is, all binary samples can be linearly represented by $r$ binary vectors. As a result, the number of unique samples is potentially related to $r$.

\section{Ordinal matrices for classification\label{sec:opt}}
In this section, we discuss the design of the optimal ordinal matrices in (\ref{eq:obj_s}). Then we discuss combining deep feature representation to improve the stability of SOM.
\subsection{The optimal ordinal matrix}
We begin the study of the optimal ordinal matrix $S$ for (\ref{eq:obj_s}) with a two-class problem. We expect that all intra-class and inter-class sample pairs of binary codes are well separated with a large margin, i.e.,
\begin{equation} \label{eq:criterion}
J(B) = {\textstyle{1 \over {\mu _1 }}}\sum\limits_{c_i  \ne c_j } {\left\| {b_i  - b_j } \right\|_0 }  - {\textstyle{1 \over {\mu _2 }}}\sum\limits_{c_i  = c_j } {\left\| {b_i  - b_j } \right\|_0 }
\end{equation}
where $B=[b_1,\dots,b_n] \in R^{m\times n}$ is a binary matrix, $\mu _1$ and $\mu _2$ are the numbers of extra-class and intra-class pairs respectively, and $\left\| . \right\|_0$ is the counting norm (i.e., the number of nonzero entries in a vector or matrix). Each row of $B^T$ corresponds to the binary code of one data item. The first term of (\ref{eq:criterion}) rewards items from difference classes having large Hamming distance, while the second term penalizes items from the same class having small Hamming distance. The maximization of $J(B)$ is NP-hard. By analyzing $J(B)$, we make the following two observations on its optimal solution,

\newtheorem{theorem}{Proposition}
\begin{theorem} \label{th:SOM_B1}
    The maximum value of $J(B)$ is equal to the number of bits ($m$), i.e., $\mathop {\max }\nolimits_B J(B) \le m$.

     {\bf Proof.} According to the definition of the $\ell_0$ norm, we can easily derive that $\mathop {\max }\nolimits_B (J(B)) < m$. In addition, when $\hat B$ satisfies,
     \begin{itemize}
\item[a)] For $\forall i,j,k$ and $ c_i \ne c_j $, if $b_{ik} \ne b_{jk}$, then ${\textstyle{1 \over {\lambda _1 }}}\sum\limits_{c_i  \ne c_j } {\left\| {b_i  - b_j } \right\|_0 }  = m$;
\item[b)] For $\forall i,j,k$ and $ c_i = c_j $, if $b_{ik} = b_{jk}$, then $ {\textstyle{1 \over {\lambda _2 }}}\sum\limits_{c_i  = c_j } {\left\| {b_i  - b_j } \right\|_0 }  = 0$,
     \end{itemize}
     we obtain $J(\hat B)=m$ (Fig. 2 (a) gives an example of $\hat B$). Hence $\mathop {\max }\nolimits_B J(B) \le m$.
\end{theorem}

\begin{theorem} \label{th:SOM_B2}
    If there exists a $\hat B$ such that $J(\hat B)= m$, the $\hat B$ satisfies the following two conditions. (a) All the samples in each class have a unique binary code. (b) The sample code of one class is orthogonal to that of the other class.
    {\bf Proof.} If $ c_i = c_j $ and $b_{ik} \ne b_{jk}$, then $\sum\limits_{c_i  = c_j } {\left\| {b_i  - b_j } \right\|_0 }  > 0$ so that $J(B)<m$. Since $b_{ik}  \in \left\{ {0,1} \right\}$ and $ \left\| {b_i  - b_j } \right\|_0  = m$ for $ c_i \ne c_j $, $b_i^T b_j  = 0$. Hence $b_i$ is orthogonal to $b_j$ when $c_i \ne c_j $ and $J(B)= m$.
\end{theorem}

\begin{figure}[t]
\centering
\subfigure[]{\includegraphics[width=26mm]{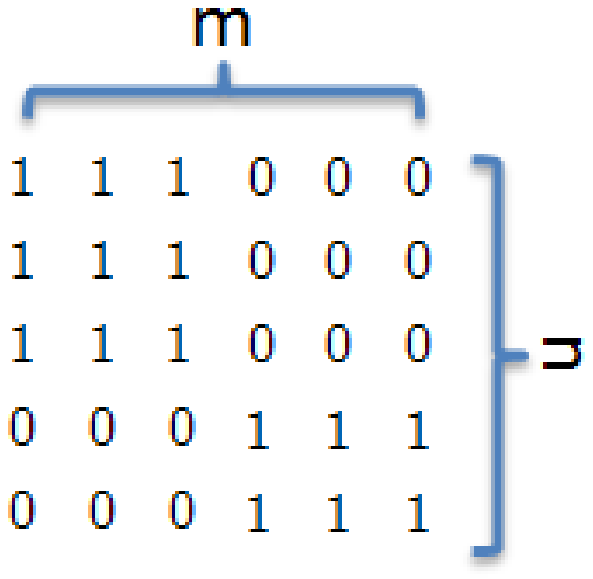}}
\subfigure[]{\includegraphics[width=27mm]{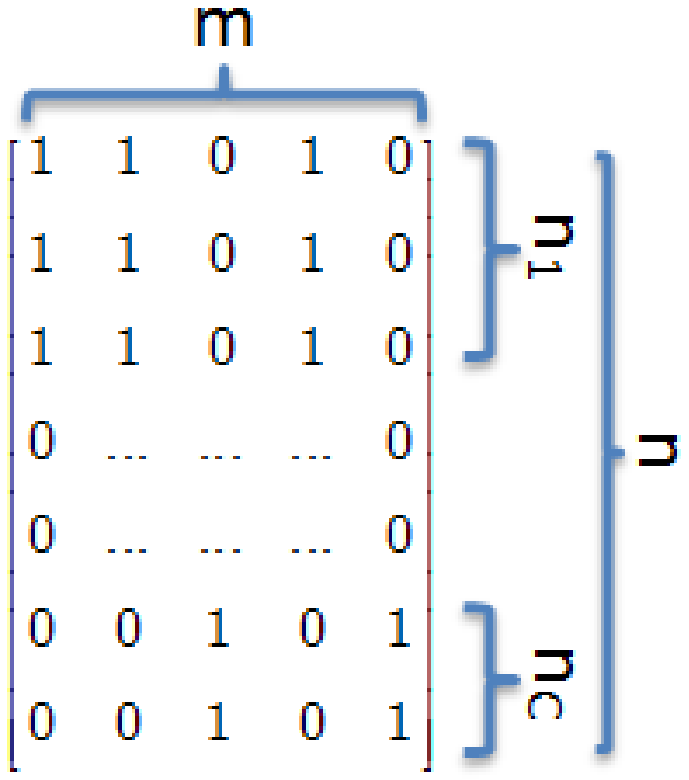}}
\subfigure[]{\includegraphics[width=27mm]{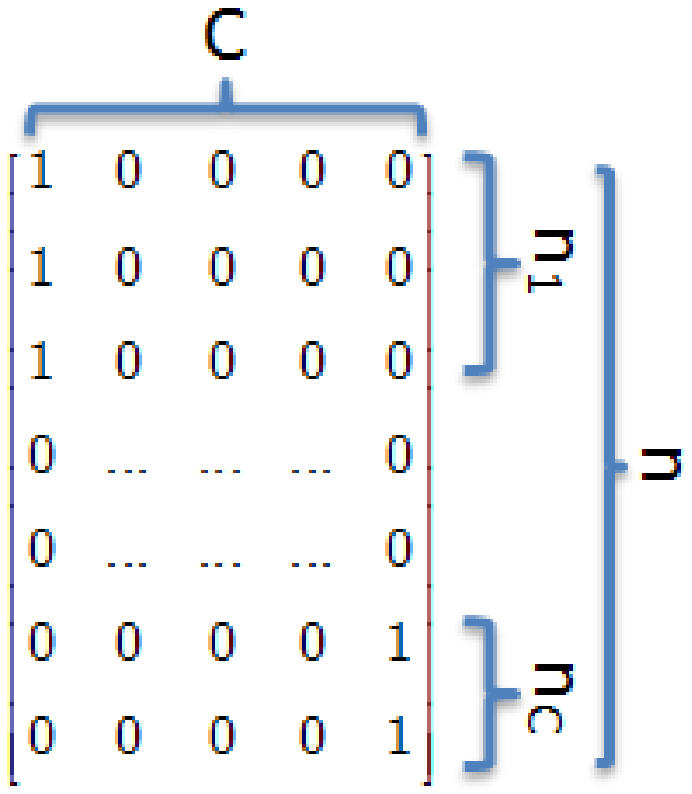}}
\caption{Three types of the optimal ordinal matrices. (a) The optimal ordinal matrix for a two-class problem. (b) Unsupervised ordinal matrix constructed via appearance information. Binary codes of all samples from the same class are arbitrary but unique and identical. (c) Supervised ordinal matrix via the spectral matrix of linear discriminant analysis~\cite{Cai:2007}.}
\label{fig:stru}
\end{figure}

From Propositions \ref{th:SOM_B1} and \ref{th:SOM_B2}, we can easily obtain the optimal ordinal matrix for a two-class problem as shown in Fig.~\ref{fig:stru} (a). Previous ordinal feature selection methods~\cite{ZSun:2009}\cite{ZSun:2014} actually select ordinal filters one by one so that the selected filters generate codes like in Fig.~\ref{fig:stru} (a). When there are multiple classes, the problem of determining the optimal binary codes becomes complex. Inspired by Propositions \ref{th:SOM_B1} and \ref{th:SOM_B2}, we consider two types of ordinal matrices to approximate the optimal ordinal matrix (shown in Fig.~\ref{fig:stru} (b)-(c)).

For the unsupervised ordinal matrix, we just require that the binary codes of each class be unique. There are many ways to generate informative binary codes for this case, e.g., random binary codes and Hadamard codes~\cite{Hedayat:1978}. Since ordinal filters perform learning based on human face appearances, we also expect that the unsupervised ordinal matrix would capture useful appearance information of video data. To accomplish this, we apply the unsupervised version of Iterative Quantization (PCA-ITQ)~\cite{YGong:2011} to the mean faces of each class to generate the corresponding unique binary code for each class. Then, the unsupervised ordinal matrix contains appearance information while the binary codes of different classes are largely uncorrelated.

For the supervised ordinal matrix, we simply employ the spectral matrix of linear discriminant analysis~\cite{Cai:2007} (the regression target of multi-class linear regression). In this spectral matrix, the binary codes of the samples from any one class have just one bit set, which define the orders of a class. Since this spectral matrix contains discriminative information, the ordinal matrix will contain supervised information if this spectral matrix is used as the ordinal matrix. However, the code length of this spectral matrix can be only $C$. If code lengths larger than $C$ are needed, we can obtain longer binary codes by combining the spectral matrix with the unsupervised ordinal matrix.

\subsection{Deep Feature Representations\label{sec:cnn}}
Since there are large variations of intra-class samples in uncontrolled VFR environments, it is often difficult to use one type of local appearance features to obtain satisfactory recognition results. Hence, biometric researchers often combine several local feature to improve generalization ability and recognition performance. In \cite{WZhang:2005}, Gabor and LBP were combined to enhance the representation power of the spatial histogram. In \cite{ZChai:2014}, Gabor ordinal measures were proposed to improve distinctiveness of Gabor features and robustness of OM's. In \cite{ZLei:2014}\cite{TChan:2014}, different techniques are combined together to achieve state-of-the-art results.

Inspired by the success of the combination of several appearance features, we couple SOM with deeply learned features from convolutional neural networks (CNN) \cite{LCun:1990} to improve coding stability. Benefiting from CNN's deep architecture and supervised learning approach~\cite{Bengio:2013}, CNN's can efficiently deal with large amounts of data and generate a hierarchical and discriminative feature representation. The use of deeply learned features makes the learned ordinal features contain not only the prior structure from data but also the hierarchical structure of local image patches.

\begin{figure*}[t]
\centering
\subfigure[the Honda/UCSD dataset]{\includegraphics[height=20mm]{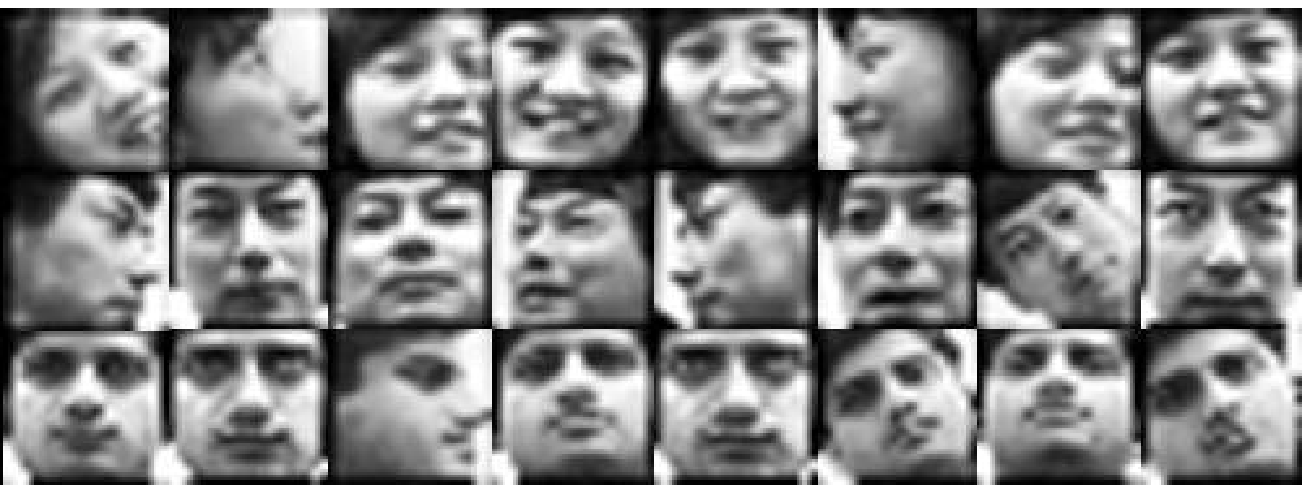}}
\subfigure[the CMU Mobo dataset]{\hspace{5mm}\includegraphics[height=20mm]{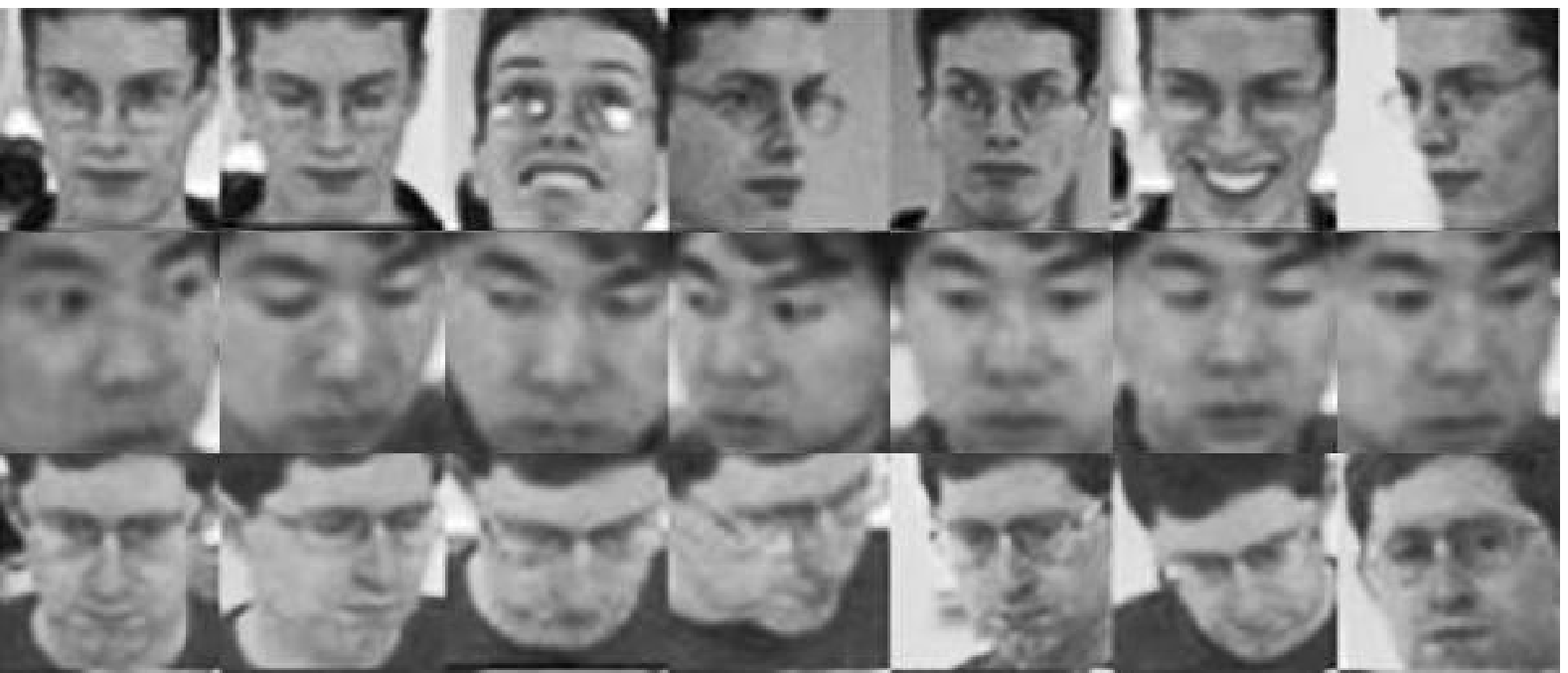}\hspace{5mm}}
\subfigure[the YouTube Celebrities dataset]{\includegraphics[height=20mm]{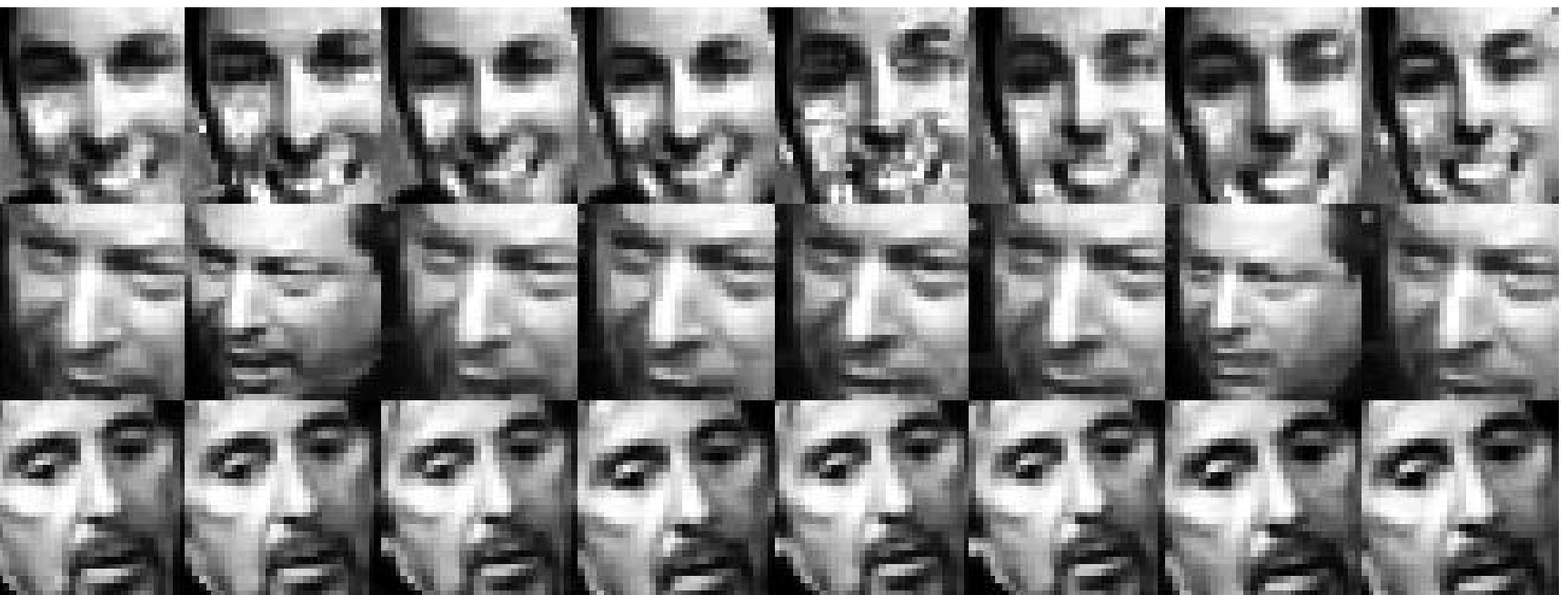}}
\caption{Cropped facial images of three different subjects in the three video databases respectively.}
\label{fig:exa}
\end{figure*}

The CNN network implemented by Alex\footnote{\url{https://code.google.com/p/cuda-convnet/}} is used as our deep architecture. This CNN first feeds gray scale images to two convolutional layers, each followed by a normalization layer and a max-pooling layer. Then, two locally connected layers are connected to the output of the second max-pooling layer, and finally to a C-way soft-max regression layer (C is the number of classes) that produces a distribution over class labels. The inputs to this network are the cropped gray scale face images without any preprocessing. The last C-way soft-max regression layer provides supervised information for learning face representations. The outputs of the last locally connected layers are employed as deep feature representations.

\section{Experiments \label{sec:exp}}
In video-sharing websites, there are a large number of face videos, each of which contains hundreds of face images. Using binary features to represent these face images will significantly save computational power and storage space. Hence, VFR is a good test platform to evaluate SOM. All experiments are run 10 times by repeating the random selection of training/testing set. For all binary code methods, the simple nearest neighbor classifier for each unique code in the probe set with voting is used as a classifier to report recognition rates.

\subsection{Methods}
We systematically compare SOM with popular techniques from three categories. {\bf SOM1} and {\bf SOM2} indicate Algorithm~\ref{Alg:SOM} using the last two structures from Fig.~\ref{fig:stru} (b)-(c) respectively. For SOM2, the bits from the optimal matrix for SOM1 is appended to that for SOM2 as discussed in Section~\ref{sec:opt} if code length is larger than the number of classes.

For the first category, we compare SOM with state-of-the-art data-driven binary feature methods in biometrics, including discriminant face descriptor (DFD)~\cite{ZLei:2014}, Gabor ordinal measures (GOM)~\cite{ZChai:2014}, and compact binary face descriptor (CBFD)~\cite{JLu:2015}. As in \cite{JLu:2015}, cosine distance is used for the three methods to achieve their best recognition accuracy. Since the feature dimensions of DFD and CBFD are too high, whitened PCA (WPCA) is applied to reduce their feature dimensions to 1000~\cite{JLu:2015}.

For the second category, we compare SOM with popular hashing methods, including locality sensitive hashing (LSH) \cite{Gionis:1999}, iterative quantization (ITQ)~\cite{YGong:2011}, kernel-based supervised hashing (KSH)~\cite{WLiu:2012}, fast supervised hashing (FastH)~\cite{GLin:2014}, and supervised discrete hashing (SDH) \cite{FShen:2015}. For ITQ, its supervised version (CCA-ITQ) and unsupervised version (PCA-ITQ) are included. PCA is used as a preprocessing step for CCA-ITQ. For SDH, we use the notation SDH-n to indicate that SDH uses image pixels rather than nonlinear RBF kernel mapping as its input. Hamming distance is computed on each pair of face samples in training/testing sets.

For the last category, we compare SOM with popular VFR methods, including discriminative canonical correlations (DCC) \cite{Kim:2007}, manifold discriminant analysis (MDA)~\cite{RWang:2008}, sparse approximated nearest point (SANP)~\cite{YHu:2011}, sparse representation for video (SRV) and its kernelized version KSRV \cite{YChen:2013}, covariance discriminative learning (Cov+PLS) \cite{RWang:2012}, jointly learning dictionary and subspace structure (JLDSS)~\cite{GZhang:2014}, image sets alignment (ImgSets)~\cite{ZCui:2012}, regularized nearest points (RNP)~\cite{MYang:2013}, and mean sequence sparse representation-based classification (MSSRC)~\cite{Ortiz:2013}. As in \cite{MYang:2013}\cite{Ortiz:2013}\cite{YChen:2013}\cite{GZhang:2014} , we directly cited the best recognition rates of these methods from the literature.

\begin{figure*}[t]
\centering
\subfigure[]{\includegraphics[height=35mm]{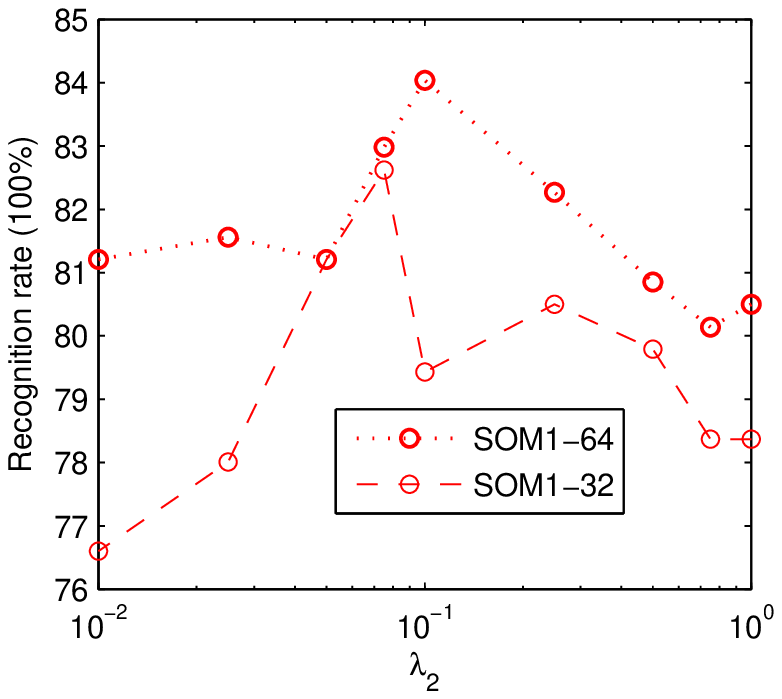}}
\subfigure[]{\includegraphics[height=35mm]{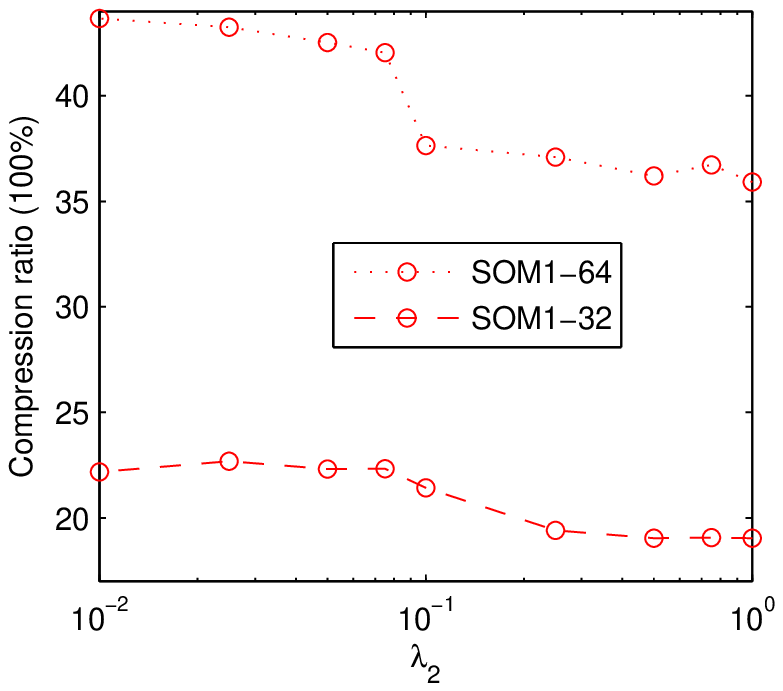}}
\subfigure[]{\includegraphics[height=35mm]{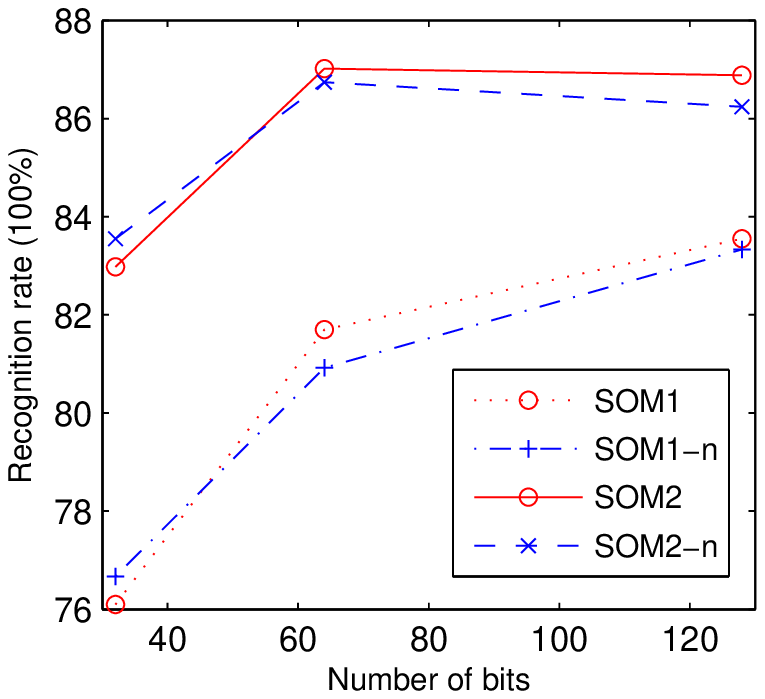}}
\subfigure[]{\includegraphics[height=35mm]{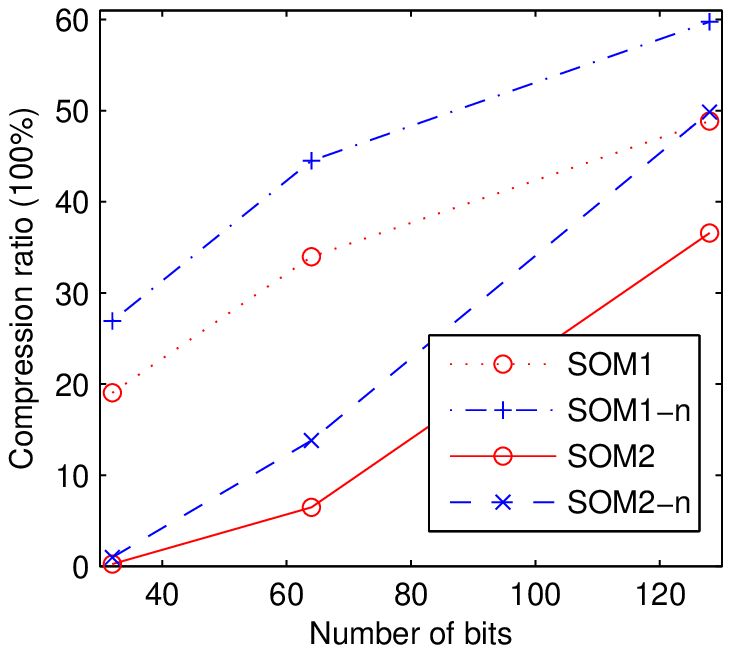}}
\caption{Recognition rates and compression ratios of SOM under different parameter setting. (a) Recognition rates as a function of $\lambda_2$. (b) Compression ratios of samples as a function of $\lambda_2$. (c) Average recognition rates with or without (\ref{eq:cor}). SOM-n indicates that the SOM method without using (\ref{eq:cor}). (d) Average compression ratios of samples with or without (\ref{eq:cor}).}
\label{fig:param}
\end{figure*}

\subsection{Databases}
Three commonly used face video datasets are used to evaluate different methods, including,

{\bf The Honda/UCSD dataset}~\cite{KLee:2003} is composed of 59 video sequences of 20 subjects. The sequences of each subject contain pose and expression variations. The lengths of the sequences vary from 12 to 645. Fig.~\ref{fig:exa} (a) shows cropped images from this dataset. We follow the standard training/testing configuration in \cite{RWang:2008}\cite{YHu:2011}\cite{RWang:2012}\cite{GZhang:2014}: 20 sequences are used for training and the remaining 39 sequences for testing. All video frames are used to report classification results. Since there are only 39 testing sequences, the improvement of recognition rates is 2.6\% (\{1/39\}*100\%) when one additional sequence is correctly classified.

{\bf The Mobo (Motion of Body) dataset}~\cite{Gross:2001} was originally published for human pose identification. It contains 96 sequences of 24 different subjects walking on a treadmill. Each subject has four video sequences corresponding to four walking patterns respectively. These patterns (slow, fast, inclined, and carrying a ball) were captured using multiple cameras. Fig.~\ref{fig:exa} (b) shows some cropped images from three subjects. We follow the standard training/testing configuration in \cite{RWang:2008}\cite{YHu:2011}\cite{RWang:2012}\cite{GZhang:2014}. One video was randomly chosen as training and the remaining three for testing. The improvement of recognition rates is (1.4\% = {1/72}*100\%) if one additional video sequence is correctly classified.

{\bf The YouTube Celebrities dataset} \cite{Kim:2008} contains 1910 video clips of 47 human subjects (actors, actresses, and politicians) from the YouTube website. Roughly 41 clips were segmented from 3 unique videos for each person. These clips are mostly low resolution and highly compressed. Each facial image is cropped to size $30 \times 30$ as shown in Fig.~\ref{fig:exa} (c). This dataset is challenging because it contains large facial variations (e.g., pose, illumination and expressions) and tracking errors in the cropped faces. Following the standard setup, the testing dataset is composed of 6 test clips, 2 from each unique video, per person. The remaining clips were used as the input to the CNN to learn a 1152-D feature representation. One frame of video (one single image) is fed into the CNN at a time. We randomly selected 3 training clips, 1 from each unique video.

\subsection{Algorithmic Analysis}
Since our SOM method consists of several parts to improve performance, we investigate the effectiveness of each part on the YouTube Celebrities dataset. To simplify parameter setting, we directly use the default parameter setting of $\mu$ and $\lambda _1$ in the LIBLINEAR SVM source code. Hence there is only one parameter $\lambda _2$ to control the effectiveness of output structures.

Fig.~\ref{fig:param} (a) and (b) show recognition rates and compression ratios of samples as a function of $\lambda _2$ respectively. Experimental results are from one single run. The lower compression ratio of an algorithm is, the better the algorithm is. We observe that parameter $\lambda _2$ affects both recognition rates and compression ratios. When $\lambda_2$ is a large, the output structure term $\left\| {B - S} \right\|_F^2$  dominates (\ref{eq:som_rank}). If $\lambda_2$ is sufficiently large, the optimal solution of $B$ will equal the ordinal matrix $S$, which indicates directly using $S$ as the class labels of SVM to perform binary code learning. When $\lambda_2$ tends to be zero, (\ref{eq:som_rank}) becomes maximum margin clustering~\cite{LXu:2004}. That is, we seek a global ordinal filter matrix $W$ to group the samples from the same class into several clusters.

Since $S^c$ is a rank-one matrix, $B$ will be a rank-one matrix if $B$ is equal to $S$. In VFR problems, a video clip often contains many face variations so that it is difficult to use one binary vector to represent all face variations. From Fig.~\ref{fig:param} (b), we also observe that the rank of the learned $B$ is larger than 1. Hence, to keep the diversity of learned $B$, it is not a good strategy to directly use $S$ as the class labels of SVM or to set $\lambda_2$ to a large value, although a larger $\lambda_2$ will result in better compression. Meanwhile, setting $\lambda_2$ too small will also damage performance. If $\lambda_2$ tends to zero, there will be no structure constraints to ensure that the learned ordinal features are similar to the optimal ordinal matrix for classification. Hence, the performance of SOM will decrease in terms of both recognition rates and compression ratios.

\begin{figure*}[t]
\centering
\subfigure[Honda]{\includegraphics[height=35mm]{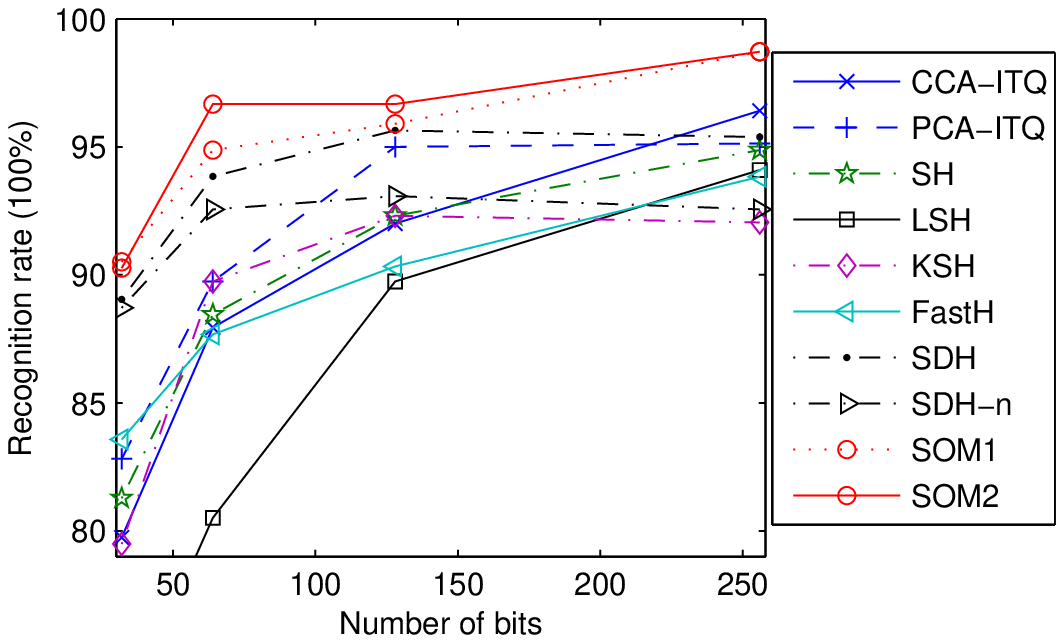}}
\subfigure[Mobo]{\includegraphics[height=35mm]{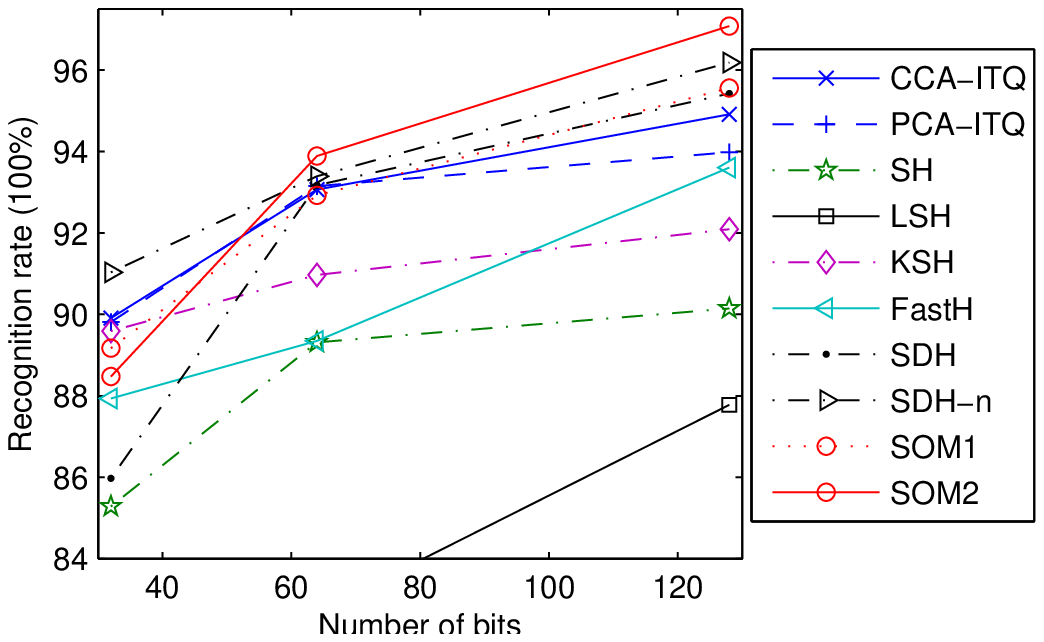}}
\subfigure[Youtube]{\includegraphics[height=35mm]{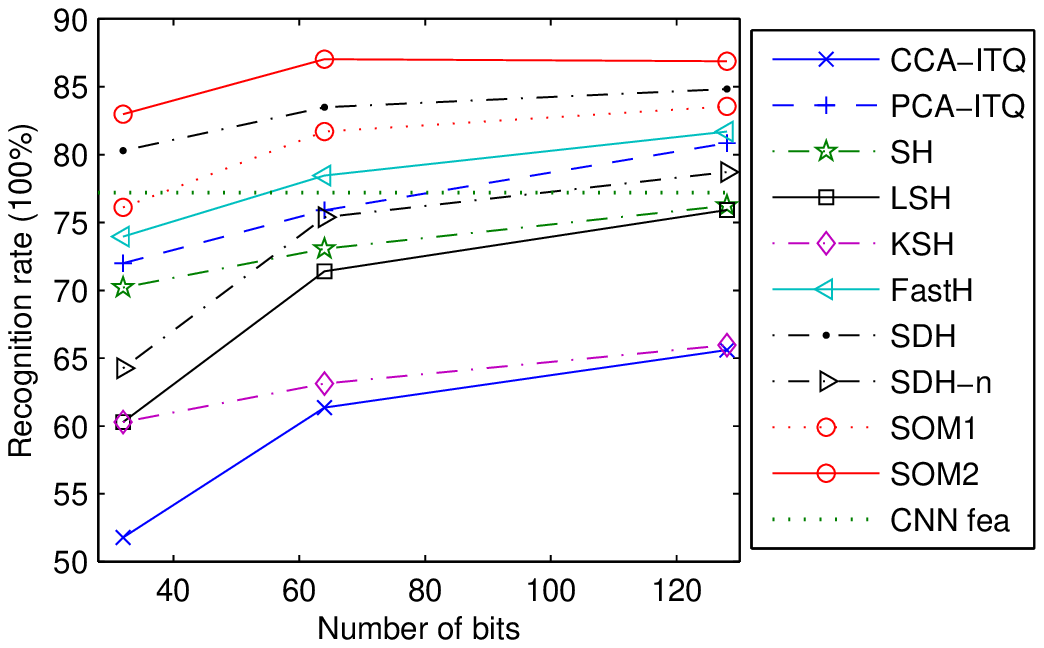}}
\caption{Recognition rates of different binary code learning methods.}
\label{fig:rate}
%\end{figure*}
%\begin{figure*}[t]
\centering
\subfigure[Honda]{\includegraphics[height=35mm]{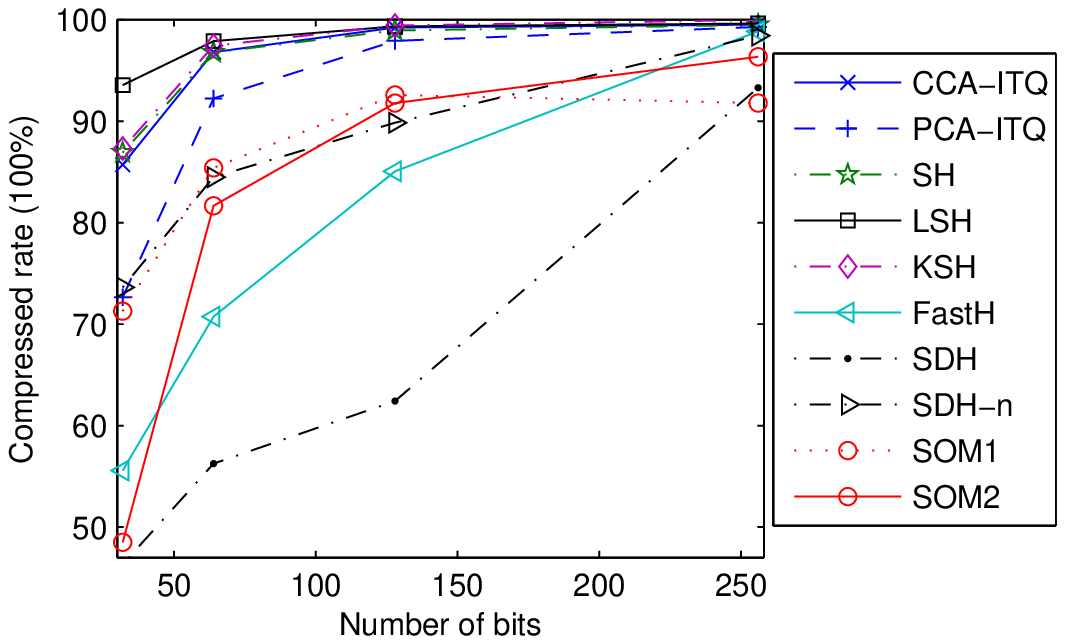}}
\subfigure[Mobo]{\includegraphics[height=35mm]{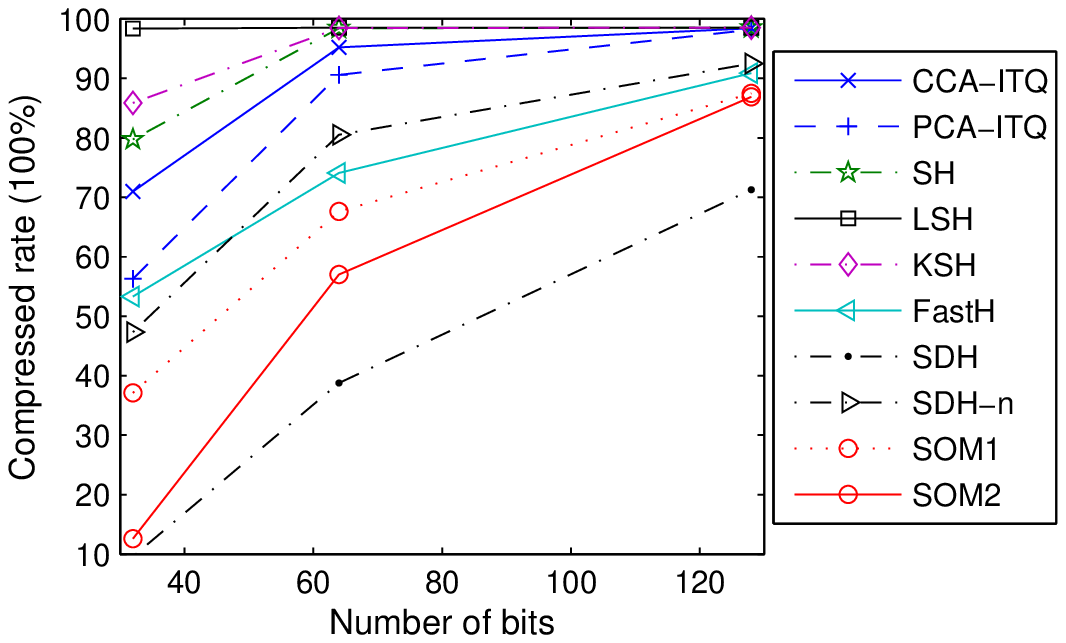}}
\subfigure[Youtube]{\includegraphics[height=35mm]{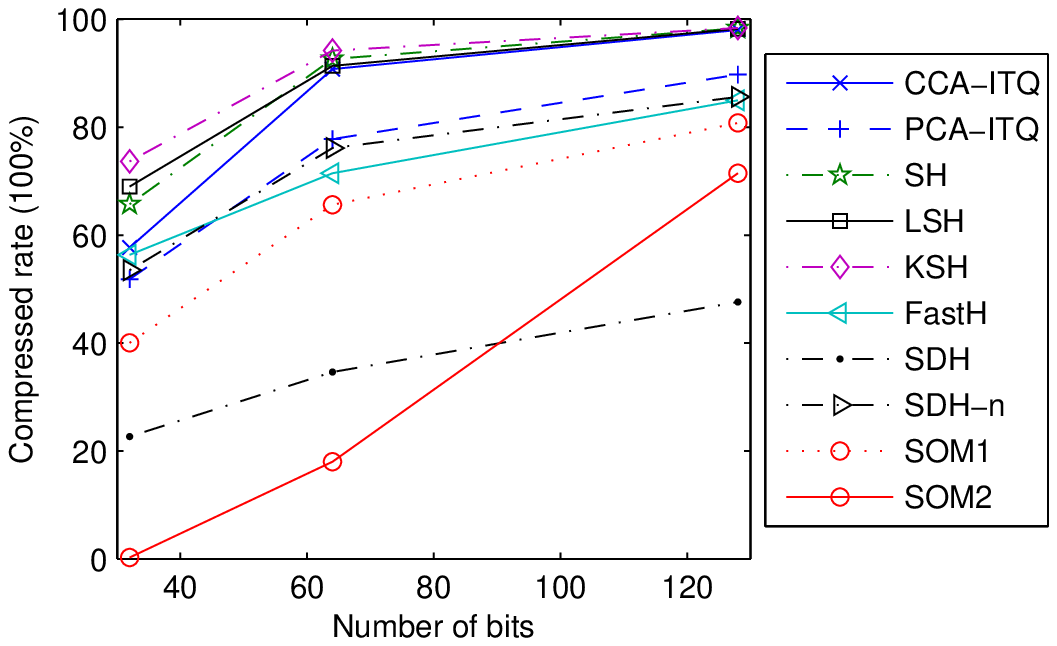}}
\caption{Compression ratios of different binary code learning methods on the three testing sets. Compression ratio = the number of unique samples/ the total number of samples. The lower compression ratio an algorithm has, the better the algorithm is.}
\label{fig:cste}
%\end{figure*}
%\begin{figure*}[t]
\centering
\subfigure[Honda]{\includegraphics[height=35mm]{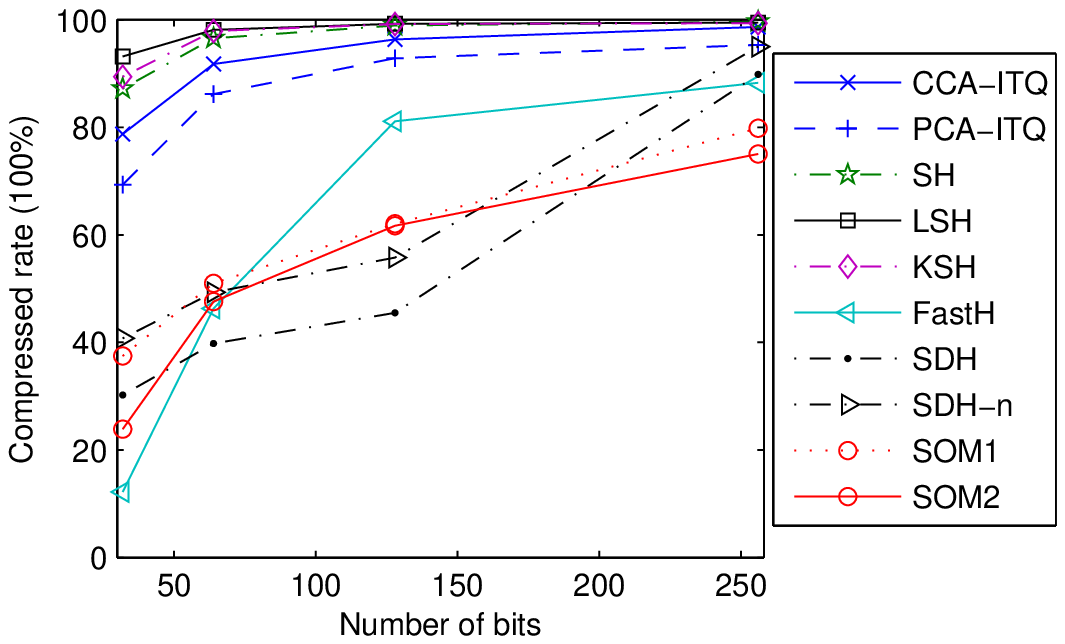}}
\subfigure[Mobo]{\includegraphics[height=35mm]{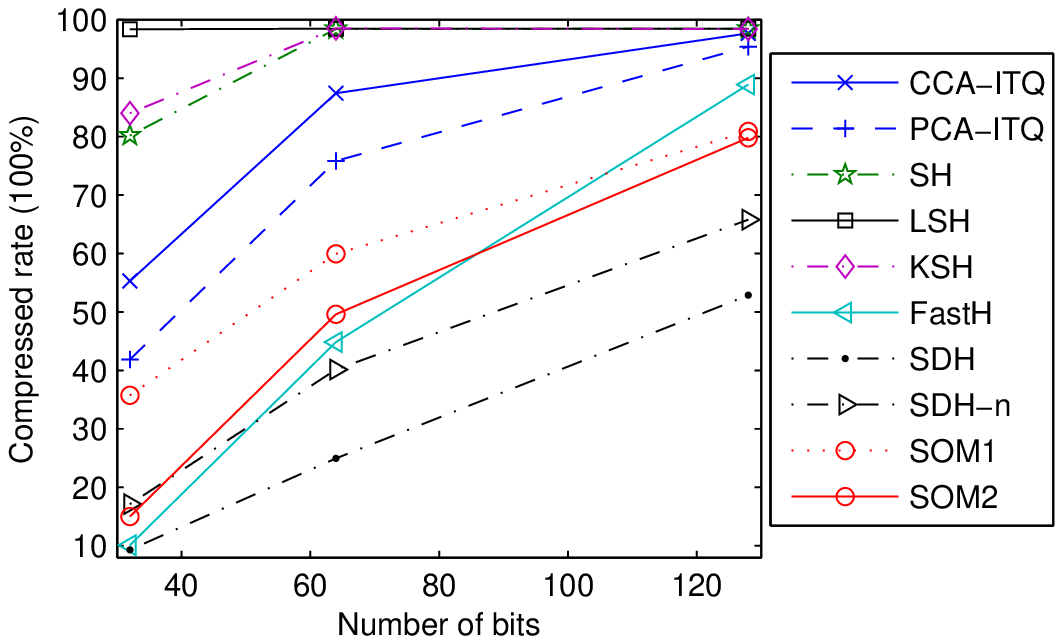}}
\subfigure[Youtube]{\includegraphics[height=35mm]{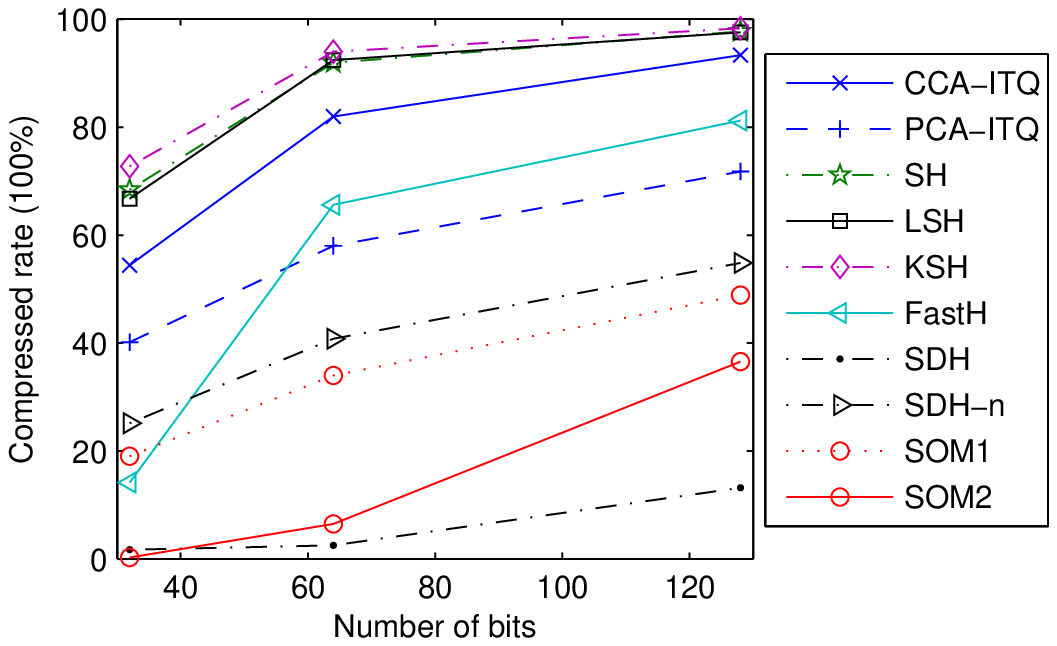}}
\caption{Compression ratios of different binary code learning methods on the three training sets.}
\label{fig:cstr}
\end{figure*}

Fig.~\ref{fig:param} (b) and (c) show recognition rates and compression ratios of samples without using (\ref{eq:cor}) respectively. SOM-n indicates that the SOM method uses $sgn(.)$ function to obtain binary codes rather than using (\ref{eq:cor}). We observe that using (\ref{eq:cor}) further improves recognition rates and reduces compression ratios. This indicates that our SOM methods can correct some binary codes such that the learned codes become correlated. Since video data often contain a large number of face samples, it is impossible to make face samples uncorrelated as assumed by hashing methods. Reducing the redundancy of video data should be helpful for performance. We also observe that the improvement using (\ref{eq:cor}) is not significant. We regard these results as reasonable because CNN features have powerful ability to learn discriminative representations. Since the binary codes learned by SOMs are discriminative enough on CNN features, there is a limited potential to further improve performance.

\subsection{Comparisons to binary code methods}
Table~\ref{tab:fea} and Figures~\ref{fig:rate},\ref{fig:cste},\ref{fig:cstr} show recognition rates and compression ratios of different binary code learning methods on the three video face databases. From these results, we make several observations:

High-dimensional and dense features are powerful for VFR. Three binary feature representation methods (GOM, CBFD and DFD) obtain the highest recognition rate (close to 100\%) on the Honda dataset, and comparable recognition rates on the other two datasets. However, the best recognition rates of these three methods are obtained by cosine distance rather than Hamming distance. Dense feature representations will result in very high computational costs for VFR. For the Honda dataset, we can see that longer codes will lead to better recognition rates. The recognition rates of CCA-ITQ, LSH, FastH, SOM1 and SOM2 increase quickly as the number of bits increases.

\begin{table*}
\centering
\begin{tabular}{|l|c|c|c|c|c|c|c|c|c|}
\hline
                  & \multicolumn{3}{c|}{Honda} & \multicolumn{3}{c|}{Mobo} & \multicolumn{3}{c|}{Youtube} \\ \hline
Methods(dim)&    RR  & CS1     & CS2     & RR     & CS1     & CS2     & RR      & CS1     & CS2      \\ \hline\hline
GOM(2560)   & 99.0\% & 100.0\% & 100.0\% & 92.6\% & 99.7\%  & 100.0\% & 68.1\%     &  99.3\% & 99.3\% \\
CBFD(32000) & 99.5\% & 99.4\%  & 100.0\% & 95.1\% & 100.0\% & 100.0\% & 66.3\% &  99.3\% & 99.3\% \\ %\hline
 DFD(50176) & 99.2\% & 100.0\% & 100.0\% & 93.6\% & 100.0\%  & 100.0\% & 64.7\%&  99.3\% & 99.3\%       \\ \hline
\end{tabular}
\caption{Experimental results of three state-of-the-art binary feature representation methods. 'RR', 'CS1' and 'CS2' indicate recognition rate, compression ratio on the testing set, and compression ratio on the training set respectively. \label{tab:fea}}
\end{table*}

Compared to the hashing methods designed for image retrieval, SOM methods are more effective for VFR. On all three databases, SOM methods achieve the highest recognition rates, and consistently outperform their hashing competitors. This may be because SOM methods can utilize and preserve the structure information from face videos. Since SOM2 considers discriminative binary codes in its prior structure, SOM2 performs better than SOM1 on the last two databases. On the YouTube database, since CNN features capture face variations well, SOM methods obtain state-of-the-art recognition rates compared to the complex classification models (e.g., image set models). It should be noted that the results for these other models are not based on CNN features, and their performance should improve if they were applied to those features. More important, SOM methods use 64-bit binary features to obtain a better result than directly using CNN features in a nearest neighbor recognition framework, which offers an impressive compression ratio of 1152-dim CNN features.

Binary code learning methods provide a potential way to reduce the number of registered samples. Since there are many face samples in a video clip, a lower compression ratio of an algorithm indicates that the algorithm needs smaller storage space and computational time. Since PCA-ITQ and CCA-ITQ aim to quantize the face samples so that they are uncorrelated, they should learn different binary codes for different samples. However, their compression ratios on the training and testing sets are smaller than 100\%. This indicates that there are some samples to have the same binary code, which makes the uncorrelated constraints work not well. In addition, compression ratios of different methods on the training set seem to be lower than those on the testing set. This indicates that there are large difference between the videos in the training and testing set so that the learned coding functions more accurately capture the facial variations in the training set than those in the testing set.

FastH, SDH, SOM1 and SOM2 obtain lower compression ratios than other methods, which indicates that these methods can reduce intra-class variations. On the Honda and Youtube databases, SDH's performance seems to mainly benefit from its nonlinear RBF kernel mapping and anchor points, which forces the data to be similar to anchor points, resulting in low compression ratios. Without the nonlinear mapping, SDHn performs no better than other methods. Since the nonlinear RBF kernel mapping is an independent step for SDH, this data mapping can also be integrated into other methods as a preprocessing step if applicable. In contrast to SDH, SOM methods employ low-rank constraints to naturally group data to different clusters (or anchor points).

The optimal ordinal matrix for classification plays an important role for SOM. Although SOM1 and SOM2 are both minimized by Algorithm~\ref{Alg:SOM}, they perform differently in terms of recognition rate and compression ratio. This is because SOM makes use of ordinal matrices as output structures that are helpful for classification. Different output structures result in different characteristic SOM's. Finding or defining the optimal ordinal matrix is still an open problem for ordinal measure and hashing. The coding theory from information theory~\cite{Hedayat:1978} may provide useful insights for binary code learning methods.

\begin{figure*}[t]
\centering
\subfigure[Honda]{\includegraphics[height=46mm]{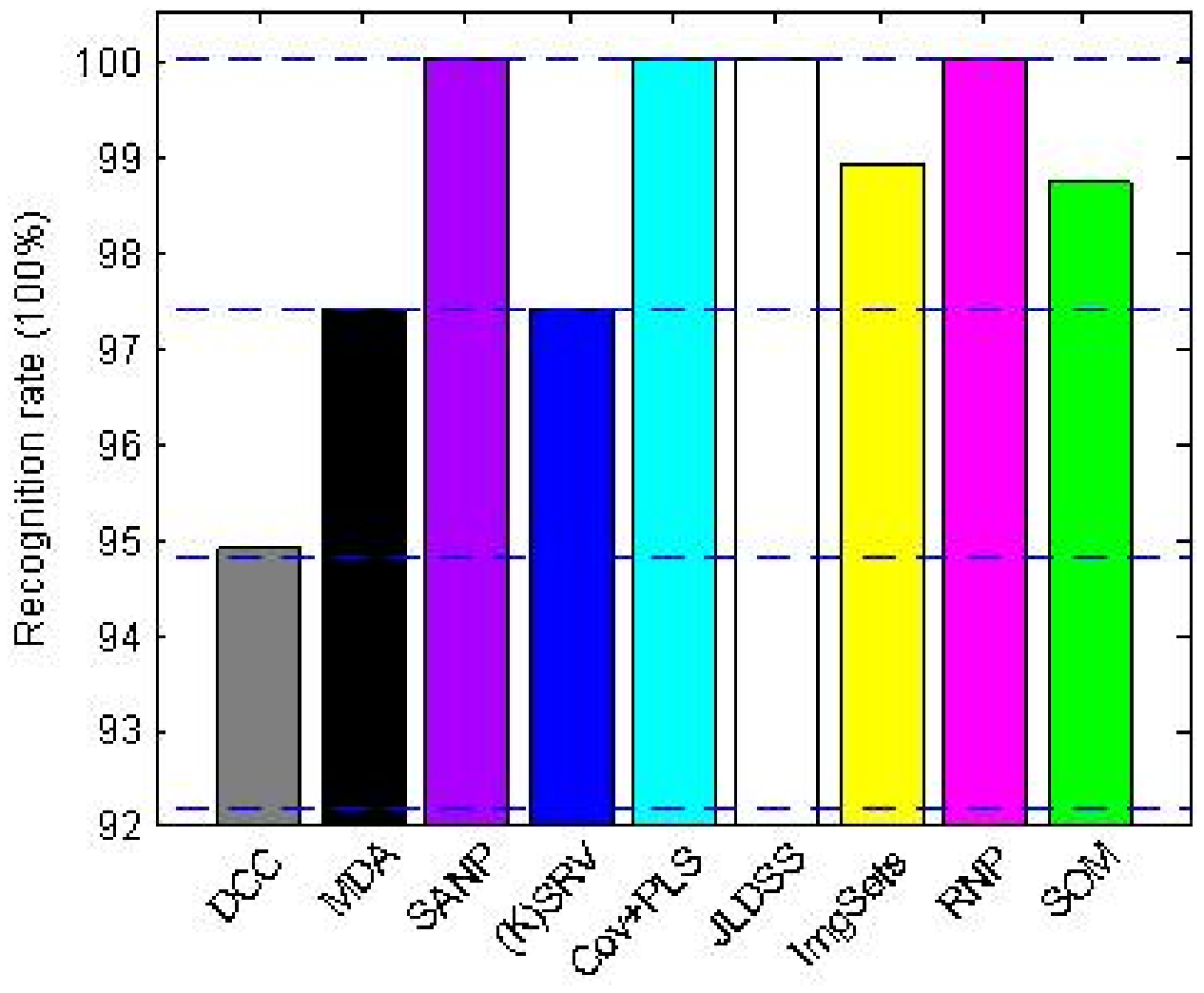}}
\subfigure[Mobo]{\includegraphics[height=46mm]{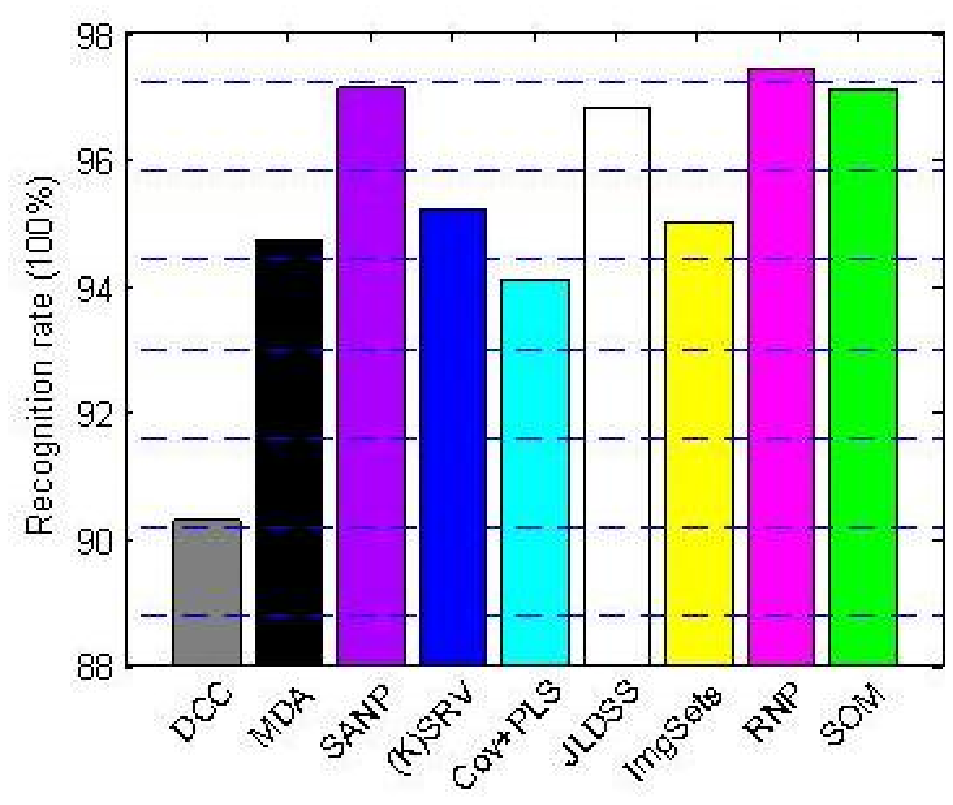}}
\subfigure[Youtube]{\includegraphics[height=46mm]{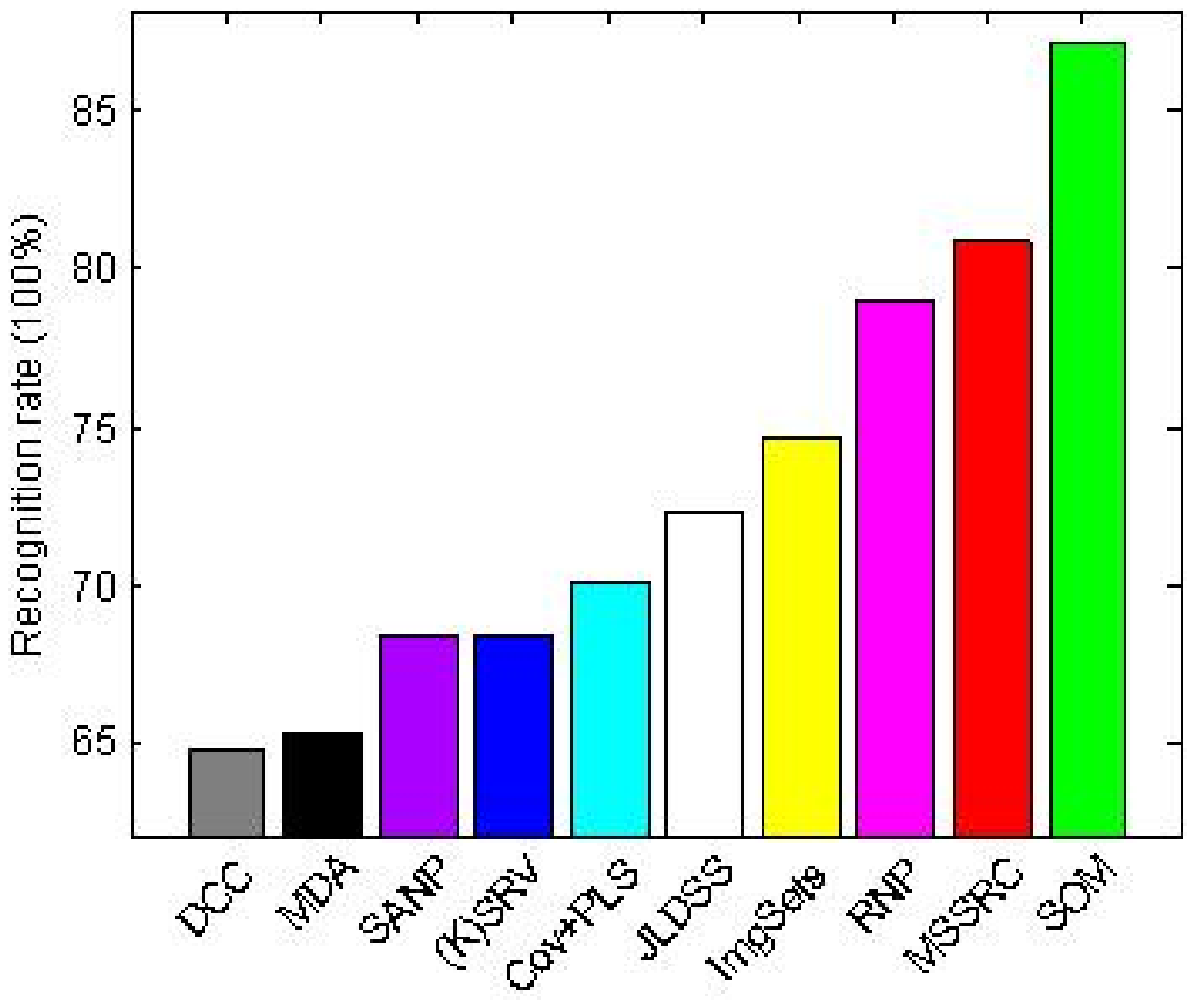}}
\caption{Recognition rates of different VFR methods on the three video databases. The interval between two dashed lines indicates the improvement of recognition rates if one additional video sequence is correctly classified.}
\label{fig:res}
\end{figure*}

\subsection{Comparisons to VFR methods}
In this subsection, we compare the proposed SOM methods with prevalent VFR methods that are based on hundreds of floating point features. Fig.~\ref{fig:res} (a) plots the average recognition rates of different VFR methods on the Honda dataset. The interval between two dashed lines indicates the improvement in recognition rates (2.6\%) if one additional video sequence is correctly classified. The highest recognition rate achieved by SOM is 98.7\% at 256 bits. We observe that the recognition rates of most of the compared methods are between 97.4\% and 100\%.  This indicates that there is at most one misclassified sequence in the randomly selected subsets. These results also show that we can use only binary features and achieve state-of-the-art results on the Honda dataset.

Fig.~\ref{fig:res} (b) plots the average recognition rates of different VFR methods on the CUM Mobo dataset. The interval between two dash lines indicates the improvement of recognition rates (1.4\% = {1/72}*100\%) if one additional video sequence is correctly classified. RNP achieves the highest recognition rate 97.4\%$\pm$1.5\%. In contrast, the recognition rate of SOM is  97.1\%. This indicates that RNP outperforms SOM in some random selection cases but not in other cases. The reason is probably that SOM simply uses a nearest neighbor classifier with voting. Since SOM is a binary feature representation method and RNP is an image set method, we consider the result of SOM to be comparable to that of state-of-the-art VFR methods. In addition, an image set algorithm can also be applied to ordinal features to further improve accuracy.

Fig.~\ref{fig:res} (c) plots the average recognition rates of different VFR methods on the Youtube dataset. We observe that MSSRC and SOM are the two best methods on this data set. Their average recognition rates are 80.8\% and 87.0\% respectively. The accuracy improvement of SOM against MSSRC is more than 6\%. The high accuracy of MSSRC is due to its robust tracker that  successfully tracked 92\% of the videos as compared to the 80\% tracked by other methods. Since the low quality of video frames incurred by the high compression rate generates large tracking errors and noise in the cropped faces \cite{YHu:2011}, a good tracker should significantly improve recognition accuracy. However, SOM did not use any preprocessing techniques (such as histogram equalization or an enhanced tracker). These results show that using a simple voting classifier can improve over the complex VFR models on the fine grained YouTube dataset. In addition, SOM can use a 64-bit representation to achieve a better recognition result than 1152-D floating point CNN representation, which offers an impressive compression ratio over CNN  features.

\section{Conclusion\label{sec:con}}
We introduced the problem of designing data-driven ordinal structures for ordinal measures learning, and developed a structured ordinal measure method for video-based face recognition. By reformulating the problem in terms of an implied equivalence relation, we posed the learning problem as a non-convex integer program problem that mainly includes two parts. The first part learns stable ordinal filters to project video data into a large-margin ordinal space. The second seeks self-correcting and discrete codes by balancing the projected data and a rank-one ordinal matrix in a structured low-rank way. Unsupervised and supervised structures are considered for the ordinal matrix. We developed an alternating minimization method to efficiently minimize the proposed non-convex formulation. Experimental results demonstrate that our SOM methods provide state-of-the-art results with fewer features and samples on three commonly used video face databases.

The future work lies in two directions. First, our results show that the proposed output structures (the optimal ordinal matrices) are useful for video-based face recognition. Hence one direction is to design or learn optimal ordinal matrix based on various facial attributes, which have been shown to further improve recognition rates. Second, our results also show that SOM can efficiently compress redundant samples, resulting in a small set of unique samples. During classification, these unique samples can be treated as representative samples or anchor points to represent all video samples. Hence another potential direction is to apply the proposed method to the area of representative sample learning.

%\section*{Acknowledgment}
%This research was supported in part by National Basic Research Program of China (grant no. 2012CB316300), National Natural Science Foundation of China (Grant nos. 61473289 and 61135002). This research is also based upon work supported in part by the Office of the Director of National Intelligence (ODNI), Intelligence Advanced Research Projects Activity (IARPA), via contract 2014-14071600010. The views and conclusions contained herein are those of the authors and should not be interpreted as necessarily representing the official policies or endorsements, either expressed or implied, of ODNI, IARPA, or the U.S. Government.  The U.S. Government is authorized to reproduce and distribute reprints for Governmental purpose notwithstanding any copyright annotation thereon.

{\small
\bibliographystyle{ieee}
\bibliography{reference}
}

\end{document}